\documentclass[10pt,twocolumn,letterpaper]{article}

\usepackage{iccv}              
\usepackage{multicol}
\usepackage{multirow}
\usepackage{makecell}

\usepackage[table]{xcolor} 

\usepackage{colortbl}

\usepackage{csquotes}
\usepackage{soul} 
\usepackage[normalem]{ulem}

\usepackage{amssymb}

\definecolor{iccvblue}{rgb}{0.21,0.49,0.74}
\usepackage[pagebackref,breaklinks,colorlinks,allcolors=iccvblue]{hyperref}

\def\paperID{11328} 
\def\confName{ICCV}
\def\confYear{2025}

\title{MixANT: Observation-dependent Memory Propagation for Stochastic Dense Action Anticipation}

\author{%
  Syed Talal Wasim$^{1,2}$ \;
  Hamid Suleman$^{1,2}$ \;
  Olga Zatsarynna$^{1,2}$ \;
  Muzammal Naseer$^{3}$ \;
  Juergen Gall$^{1,2}$
  \vspace{0.5em} \\
  $^{1}$University of Bonn \; 
  $^{2}$Lamarr Institute of ML and AI \;
  $^{3}$Khalifa University
}

\begin{document}
\maketitle
\begin{abstract}

We present MixANT, a novel architecture for stochastic long-term dense anticipation of human activities. While recent State Space Models (SSMs) like Mamba have shown promise through input-dependent selectivity on three key parameters, the critical forget-gate ($\textbf{A}$ matrix) controlling temporal memory remains static. We address this limitation by introducing a mixture of experts approach that dynamically selects contextually relevant $\textbf{A}$ matrices based on input features, enhancing representational capacity without sacrificing computational efficiency. Extensive experiments on the 50Salads, Breakfast, and Assembly101 datasets demonstrate that MixANT consistently outperforms state-of-the-art methods across all evaluation settings. Our results highlight the importance of input-dependent forget-gate mechanisms for reliable prediction of human behavior in diverse real-world scenarios. The project page is available at \url{https://talalwasim.github.io/MixANT/}. 

\end{abstract}    
\section{Introduction}
\label{sec:intro}

Human activity anticipation represents a fundamental challenge in computer vision, requiring models to predict future human actions from partial observations. This capability is crucial for applications ranging from assistive technologies to autonomous driving systems, where timely intervention depends on accurate forecasting of human intentions and behaviors. In particular, long-term dense action anticipation~\cite{gong2022future, farha2020gcpr, Farha_2018_CVPR} presents a particularly challenging problem: generating multiple and continuous, frame-by-frame predictions of future activities, often extending several minutes ahead. In order to handle the uncertainty for this task, stochastic approaches have been proposed~\cite{farha2019uaaa,zhong2023diffant,zatsarynna2024gtd,zatsarynna2025manta} that predict multiple plausible future samples. While recent approaches~\cite{zhong2023diffant,zatsarynna2024gtd,zatsarynna2025manta} use a diffusion model to generate multiple predictions for the same observations, these approaches differ in the architecture that is used within the diffusion model, including gated temporal convolutions~\cite{zatsarynna2024gtd} and Transformers~\cite{zhong2023diffant}. While Transformers have shown impressive capabilities in modeling long-range dependencies, their quadratic computational complexity with respect to sequence length limits their practical application to extended temporal horizons as they occur in long-term dense anticipation. MANTA~\cite{zatsarynna2025manta} thus utilizes the Mamba architecture~\cite{gu2023mamba} due to its remarkable capabilities in handling long sequences with near-linear computational complexity. 

\begin{figure}[t]
    \raggedright
        \includegraphics[width=\columnwidth]{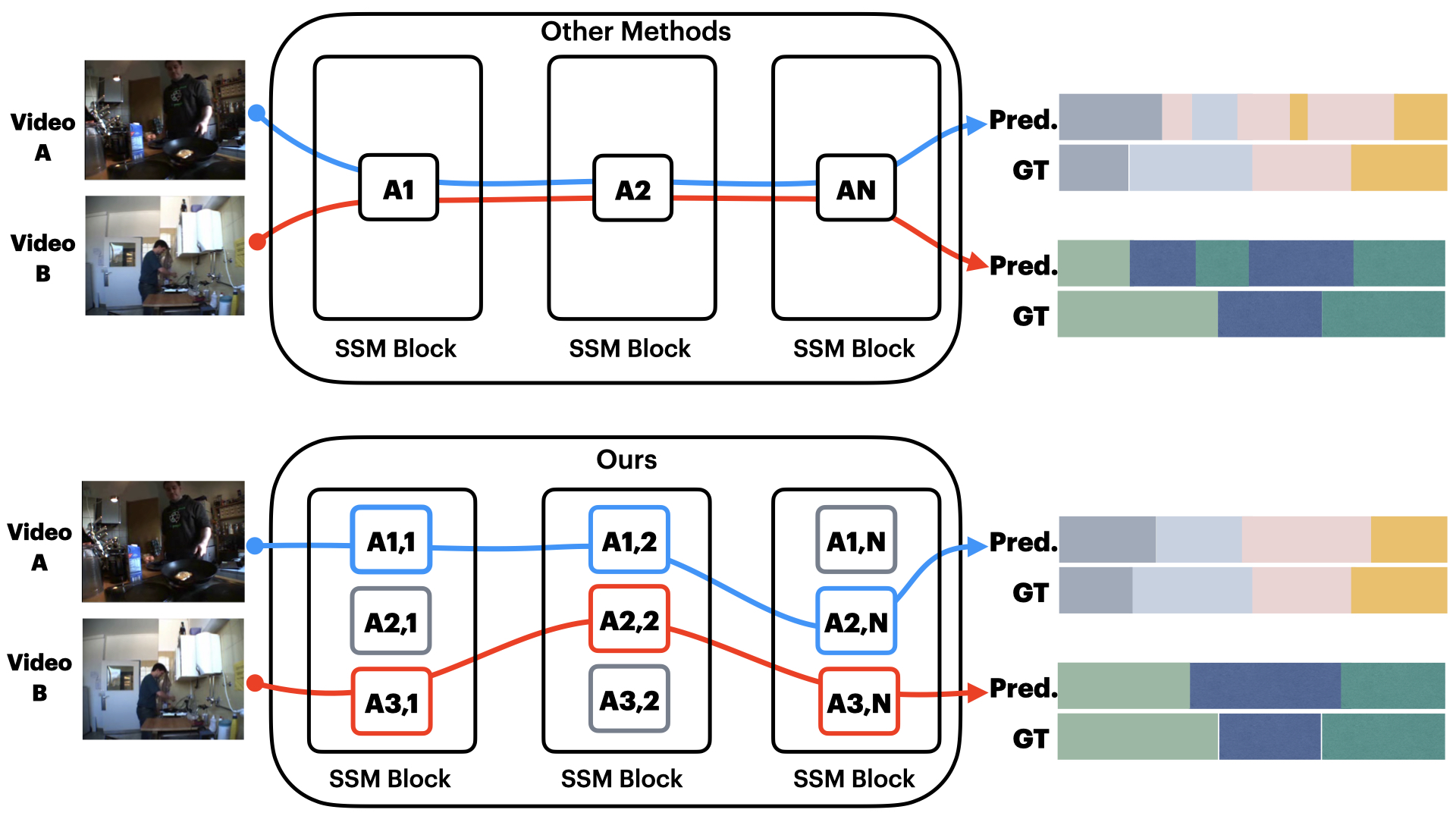}
        \captionsetup{width=\columnwidth}
        \caption{The current state-of-the-art method~\cite{zatsarynna2025manta} employs standard Mamba blocks for stochastic dense action anticipation. In contrast, we introduce the MixMamba block, where each block dynamically selects an appropriate $\textbf{A}$ matrix based on the input sequence.}
        \label{fig:teaser}
    \label{fig:teaser_combined}
    \vspace{-0.25cm}
\end{figure}

\begin{figure*}[t!]
    \centering
    \includegraphics[width=\textwidth]{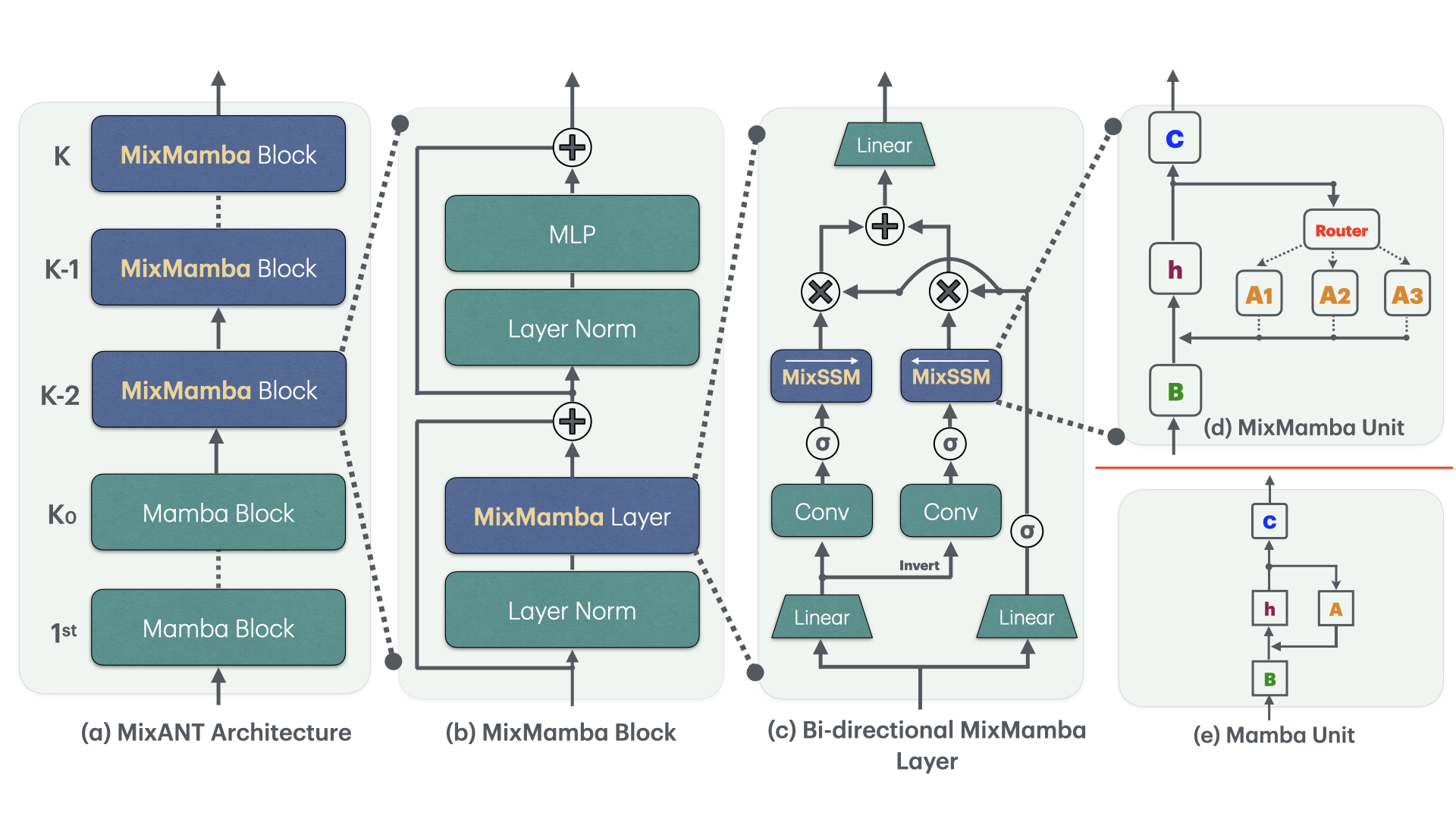}
    \\
    \vspace{-0.7cm}
    \captionsetup{width=\textwidth}
    \caption{Our architecture and its components. In the MixANT architecture (a), the first $K_0$ blocks are standard Mamba blocks, and the rest ($K-K_0$) are MixMamba blocks (b). Each MixMamba block contains a bi-directional MixMamba layer (c). In contrast to the standard Mamba unit (e), where $\textbf{A}$ is fixed for all input sequences, the MixMamba unit (d) contains a router that selects the relevant $\textbf{A}$ matrix depending on the input sequence.}
    \label{fig:main}
    \vspace{-0.2cm}
\end{figure*}

The standard Mamba model~\cite{gu2023mamba} applies input-dependent selectivity to three of its four key parameters, dynamically adapting the model's behavior based on contextual cues within the observed sequence. Specifically, these parameters include the input gate, output gate, and time-step parameter. However, one critical parameter—the forget-gate (the $\textbf{A}$ matrix) that controls how the hidden state evolves over time, lacks this input dependence. This forget-gate parameter determines how much past information is remembered or forgotten. While \cite{zatsarynna2025manta} demonstrated that input dependence on the input, output, and time-step parameters is particularly important for long-term dense action anticipation, the fourth parameter, the matrix $\textbf{A}$, is independent of the input as illustrated in \cref{fig:teaser_combined}.

This raises a particularly important research question, as the $\textbf{A}$ matrix's role in controlling temporal memory is particularly relevant for anticipation tasks. 
Depending on the specific input context, sometimes past information is critically important, while in other contexts, it may be less relevant. Moreover, in the stochastic long-term dense anticipation task, the input sequence consists of observed data and padded zeros representing future unknown labels. In this scenario, it is also desirable to have selectivity on the hidden state through forget-gates such that it selectively ignores the padded zero sequence. Therefore, any form of input dependence on this matrix, which controls the evolution of the hidden state, is important. However, introducing input dependence to the $\textbf{A}$ matrix presents substantial technical challenges. Direct computation would either require query-key like multiplication, which diminishes the sub-quadratic complexity advantage of Mamba models, or a large MLP, which adds significant computational overhead.

In this work, we address this challenge by introducing MixANT, a mixture of $\textbf{A}$ for ANTicipation, which enhances the network's representational capacity with minimal computational overhead. As shown in \cref{fig:main}, our method employs layers called MixMamba that dynamically select the most contextually relevant $\textbf{A}$ matrix based on input features. This approach introduces input dependence without sacrificing computational efficiency. To the best of our knowledge, we propose the first mixture of experts approach specifically for the $\textbf{A}$ matrix in Mamba-based architectures.

We demonstrate that this novel architecture achieves state-of-the-art performance on the stochastic long-term dense anticipation task through extensive experimentation on three benchmark datasets: 50Salads~\cite{Stein2013_50Salads}, Breakfast~\cite{Kuehne2014_breakfast}, and Assembly101 \cite{sener2022_Assembly101}. Our approach consistently outperforms existing methods across all evaluation settings, and our results highlight the potential of our approach 
to advance the field of stochastic dense action anticipation and enable reliable prediction of human behavior in diverse real-world scenarios.

\section{Related Work}
\label{sec:related_work}

Action anticipation research \cite{zhong2023surveydeeplearningtechniques} is divided into different branches, but mainly consists of short-term approaches (predicting actions within seconds)~\cite{zatsarynna2023goal, zatsarynna2021multi, girdhar2021_AVT, sener2020temporal, furnari2020rulstm, zhao2023_antgpt, mittal2024_omlette, zhong2023anticipative, zhao2022real}
and long-term approaches (forecasting multiple actions over minutes). Our work focuses on long-term dense anticipation, which requires predicting both action sequences and their durations across specified future frames. Unlike methods that treat anticipation as ordered~\cite{mascaro2023intention,das2022video_plus, zhao2023_antgpt, mittal2024_omlette} or unordered \cite{zhao2023_antgpt, zhong2023diffant, nawhal2022_rethinking}  transcript~prediction without temporal boundaries, our work focuses on long-term dense anticipation. This application requires predicting both action sequences and their durations across specified future frames. 

Dense anticipation approaches further split into deterministic models~\cite{gong2022future, Farha_2018_CVPR, Ke_2019_CVPR, sener2020temporal} that generate single predictions, and stochastic methods~\cite{farha2019uaaa, zhao2020async} — including ours — that produce multiple plausible futures for each observation. Other methods, e.g.~\cite{zatsarynna2024gtd, zhong2023diffant}, can generate predictions in both stochastic and deterministic settings. \cite{farha2019uaaa} pioneered stochastic dense anticipation by extending RNN architecture to generate multiple samples. \cite{zhao2020async} enhanced this framework using adversarial learning to improve prediction diversity. Recent approaches \cite{zatsarynna2024gtd, zhong2023diffant} have advanced beyond requiring ground-truth actions by performing simultaneous past classification and future forecasting. Specifically, \cite{zhong2023diffant} combined a diffusion model with the FUTR encoder~\cite{gong2022future} to model uncertainty in future actions, while~\cite{zatsarynna2024gtd} introduced a gated temporal diffusion network that models uncertainty in both past and future actions, achieving state-of-the-art results.

Traditional attention-based architectures face efficiency challenges due to quadratic complexity with sequence length, inspiring the development of State Space Models~\cite{gu2022efficientlyS4, smith2023simplifiedS6} that offer linear scaling with sequence length. Mamba~\cite{gu2023mamba} enhances this framework through selective state space modeling with input-dependent selection mechanisms, making it particularly effective for long sequence processing. This efficiency has led to Mamba's adaptation across various fields, including anticipation \cite{zatsarynna2025manta}, vision~\cite{liu2024VMamba, li2025VideoMamba}, and as backbones for diffusion models in generative tasks~\cite{hu2024zigma, mo2024scaling}. Recently, \cite{zatsarynna2025manta} also utilized bi-directional Mamba blocks similar to \cite{li2025VideoMamba} for stochastic long-term anticipation.
Our work uniquely leverages Mamba's capabilities by introducing a novel ``mixture of $\textbf{A}$" matrices approach that addresses the shortcomings of~\cite{gu2023mamba} mentioned in the introduction. 

The Mixture of Experts (MoE) architectures allow for significant scaling of model parameters with minimal computational overhead. First proposed by~\cite{jordan1994moe_orig, jacobs1991_moe_orig}, MoE leverages conditional computation, i.e., activating only specialized ``expert" sub-networks for specific inputs. By implementing this sparse activation strategy, MoE-based models can substantially increase model parameters with reasonable computational demands. 
The MoE technique has been successfully applied to both transformer-based and, more recently, Mamba-based architectures. Similar to our architecture in spirit are Mamba-based methods leveraging a mixture of experts (MoE)~\cite{anthony2024blackmamba, pioro2024moe_mamba, liang2025mixture_of_mamba}. However, these approaches differ from our method in key ways. Both~\cite{anthony2024blackmamba} and~\cite{pioro2024moe_mamba} utilize standard Mamba blocks~\cite{gu2023mamba} while employing MoE only to MLPs outside the Mamba block. Meanwhile, \cite{liang2025mixture_of_mamba} employs a mixture-based architecture for modality-aware selection, but unlike our method,d it uses a shared $\textbf{A}$ matrix (of Mamba block) constraining the expressivity of the network.
\section{Observation-dependent Memory Propagation in Mamba}
\label{sec:mixture_a_mamba}

The work~\cite{zatsarynna2025manta} showed that state-space models outperform transformer and other architectures for stochastic long-term dense action anticipation. The approach utilizes the bi-directional Mamba layer~\cite{gu2023mamba,li2025VideoMamba} for modeling temporal relations between past observations and future actions. A key feature of this block compared to previous state-space models is the input-dependent parameters, as we will discuss in \cref{sec:mamba_layer}. The temporal memory propagation, however, which is steered by the matrix $\textbf{A}$, remains input-independent as it is illustrated in \cref{fig:teaser}, \ie, the learned matrices $\textbf{A}$ for each block do not change for different observations. This poses a major limitation for action anticipation since the way temporal memory should be propagated between observations and future actions depends on the context of the observation. Indeed, we will show that the matrices $\textbf{A}$ vary depending on the semantic context, see \cref{fig:tsne}.

We will first describe the Mamba layer~\cite{gu2023mamba,li2025VideoMamba} in \cref{sec:mamba_layer} and our extension in \cref{sec:mixmamba_layer}. The entire approach for stochastic long-term dense action anticipation is then described in \cref{sec:method}.

\subsection{Mamba Layer}
\label{sec:mamba_layer}

State-space models define a mapping from an input sequence $x_t$ to an output sequence $y_t$ through a latent state $h_t$ using a first-order ordinary differential equation 
discretized with a time-step parameter $\Delta$:
\begin{align}
    \label{eq:tr_1}
    h_t &= \bar{\textbf{A}} h_{t-1} + \bar{\textbf{B}} x_t, \\
    \label{eq:tr_2}
    y_t &= \textbf{C} h_t, \\
    \label{eq:discr_1}
    \bar{\textbf{A}} &= \exp(\Delta \textbf{A}), \\
    \label{eq:discr_2}
    \bar{\textbf{B}} &= (\Delta \textbf{A})^{-1} (\exp(\Delta \textbf{A}) - I) \Delta \textbf{B}.
\end{align}

Mamba \cite{gu2023mamba} enhances this framework by making most parameters input-dependent, \ie, $\textbf{B}(x) \in \mathbb{R}^{B \times T \times N}$, $\textbf{C}(x) \in \mathbb{R}^{B \times T \times N}$, and $\Delta(x) \in \mathbb{R}^{B \times T \times D}$ are computed as functions of the input $x \in \mathbb{R}^{B \times T \times D}$ where $B$ is batch-size, $T$ is sequence length, $N$ is latent state dimension, and $D$ is channel dimension. $\Delta$ controls the flow of information from the input $x_t$ to the hidden state $h_t$. $\bar{\textbf{B}}$ modulates the input. $\bar{\textbf{A}}$ serves as a \textit{forget-gate}, determining which information persists from previous states and how quickly past information decays, effectively controlling the model's memory duration. Finally, the $\textbf{C}$ parameter filters state information for outputs, determining which aspects of the internal state are expressed in the output signal. While \cite{gu2023mamba} made the hypothesis that it is sufficient to make $\Delta$ input-dependent and keep $\textbf{A}$ input-independent, we demonstrate that this is not true for the task of long-term dense action anticipation, where very long sequences are processed.           

\subsection{MixMamba Layer}
\label{sec:mixmamba_layer}

Making the $\textbf{A}$ matrix input-dependent presents substantial computational challenges. A straightforward implementation would either require query-key multiplication, which would negate Mamba's sub-quadratic complexity advantage for long sequences, or necessitate a very large MLP, which would introduce significant computational overhead and parameter inefficiency. For dense anticipation tasks specifically, where models must effectively reason over both observed and unobserved regions of input sequences, this limitation is particularly restrictive. An input-dependent $\textbf{A}$ matrix would enable more nuanced control over how past information is selectively retained or forgotten based on the content of observed tokens.

To address these challenges while maintaining computational efficiency, we propose a mixture-of-experts approach for the $\textbf{A}$ matrix. Rather than computing a single input-dependent $\textbf{A}$ matrix directly, we maintain multiple expert $\textbf{A}$ matrices and employ a softmax gating mechanism to select the appropriate combination based on input characteristics.

Building upon the S6 algorithm~\cite{gu2023mamba}, we propose S6$^{+}$, which enhances the selection mechanism by implementing a mixture approach for the $\textbf{A}$ matrix. Instead of using a single fixed $\textbf{A}$ matrix, we define a set of $E$ expert matrices $\{\textbf{A}_1, \textbf{A}_2, ..., \textbf{A}_E\} \in \mathbb{R}^{E \times D \times N}$. For a given input sequence, we compute a gating vector $\gamma(x) \in \mathbb{R}^{B \times E}$ based on the projected mean of the input tokens $\gamma$:
\begin{align}
    \label{eq:gating_vector}
    \gamma(x) = \text{softmax}(W_g \cdot \text{mean}(x)),
\end{align}
where $W_g \in \mathbb{R}^{D \times E}$ is a learnable projection matrix. 
The matrix $\textbf{A}$ is then selected by:
\begin{align}
    \label{eq:mixture_a}
    \textbf{A}(x) = \textbf{A}_{\hat{e}},\quad \hat{e}= \arg\max_e \gamma_e(x),
\end{align}
where $\gamma_e(x)$ is the $e$-th element of the routing vector $\gamma(x)$. Note that during training, both $\{\textbf{A}_1, \textbf{A}_2, ..., \textbf{A}_E\}$ and $\gamma(x)$ are learned, but it requires an additional loss term as we will discuss in \cref{sec:training_inference}.

The general structure of our MixMamba layer follows the bi-directional approach~\cite{li2025VideoMamba} with the S6$^{+}$ algorithm as illustrated in \cref{fig:main}(c). The processing flow for both paths is given by
\begin{align}
    \hat{F}_{\text{Fwd}} &= \sigma(H^{K} * \text{FF}(F_t^{l})), \quad
    \hat{F}_{\text{Bwd}} = \sigma(H^{K} * \text{FF}(\overleftarrow{F_t^{l}})),
\end{align}
where $H^{K}$ is a 1D convolutional filter with size $K$, flipping across the temporal dimension is denoted as $\leftarrow$, and $*$ represents the convolution operator. For both forward and backward paths, we apply our mixture-of-experts approach:
\begin{align}
    W_t^{l} &= \text{S6}^{+}(\hat{F}_{\text{Fwd}}), \quad
    B_t^{l} = \text{S6}^{+}(\hat{F}_{\text{Bwd}}).
\end{align}
Similar to the Mamba layer, we apply a residual gating mechanism and combine the outputs:
\begin{align}
    W_t^{l} &= W_t^{l} \odot R_t^{l}, \quad 
    B_t^{l} = \overleftarrow{B_t^{l}} \odot R_t^{l}, \\
    O_t^{l} &= \text{FF}(W_t^{l} + B_t^{l})
\end{align}
where $R_t^{l} = \sigma(\text{FF}(F_t^{l}))$ and $\odot$ represents element-wise multiplication.

This mixture-of-experts approach for the $\textbf{A}$ matrix enhances the model's ability to selectively process temporal information, which is crucial for tasks involving long sequences with varying temporal dependencies. By dynamically weighting different expert $\textbf{A}$ matrices based on input characteristics, the S6$^{+}$ algorithm can adapt its temporal processing strategy to the specific patterns present in the data, enabling more effective modeling of complex sequential relationships.
\section{MixANT Architecture for Stochastic Long-Term Dense Action Anticipation}
\label{sec:method}

In this section, we present our approach for long-term stochastic dense action anticipation that is based on the bi-directional MixMamba layer described above in \cref{sec:mixmamba_layer}. For a fair comparison, we follow the diffusion architecture \cite{zatsarynna2025manta}, which we discuss in \cref{sec:diffusion}. In \cref{sec:mixant_architecture}, we describe the proposed MixANT architecture, which is illustrated in \cref{fig:main}(a). In \cref{sec:training_inference}, we finally describe the loss functions. Our final loss includes a novel load balancing loss, which encourages that the router in the MixMamba unit (\cref{fig:main}(d)) utilizes and learns all matrices and not only a subset during training.

\subsection{Diffusion for Stochastic Long-Term Dense Action Anticipation}
\label{sec:diffusion}

Following \cite{zatsarynna2024gtd}, we employ a diffusion model framework for stochastic long-term dense action anticipation. The diffusion model learns the distribution of per-frame actions given a set of observed frames:  
\begin{align}
\tilde{Y}_{1:P+F} \sim p(Y_{1:P+F} \mid \mathbf{x}_{1:P}),
\end{align}
where $P$ is the number of observed frames, $F$ is the number of future frames to predict, $\mathbf{x}_{1:P}$ is the sequence of observed frames, $Y_{1:P+F}$ represents the random variables for each frame of the entire sequence, and $\tilde{Y}_{1:P+F}$ represents the sampled action label for each frame. To condition the anticipation on the observed visual frames, a conditioning vector $\mathcal{X}$ is created by extending the observed visual features with zeros for future frames:
\begin{align}
\label{eq:padding}
\mathcal{X} = \{ \phi(x_1), \ldots, \phi(x_P), \underset{F}{\underbrace{0, \ldots, 0}}\}
\end{align}
where $\phi(x_i)$ represents visual features extracted from the $i$-th observed frame. For a fair comparison, we use the same features that have been used in previous works. After sampling $\hat{Y}_{T} \sim \mathcal{N}(\mathbf{0}, \mathbf{I})$, the diffusion model gradually generates a sample $\tilde{Y}_{1:P+F}=\hat{Y}_{0}$  after $T$ iterations
\begin{equation}
\hat{Y}_{t-1} = G_{\theta}(\hat{Y}_{t}, \mathcal{X}, t),
\end{equation}
based on an anticipation generator $G_{\theta}$ with learnable parameters $\theta$. The index $t$ denotes the diffusion step. While~\cite{zatsarynna2024gtd} uses gated temporal convolutions and~\cite{zatsarynna2025manta} a Mamba architecture for the generator $G_{\theta}$, we harness our proposed MixANT architecture, which we will describe in \cref{sec:mixant_architecture}. During inference, multiple predictions are sampled using DDIM \cite{song2020denoising} for a single set of input frames $\mathbf{x}_{1:P}$. For more details regarding the diffusion model, we refer the reader to~\cite{zatsarynna2024gtd}. 

\subsection{MixANT Architecture}
\label{sec:mixant_architecture}

Our proposed MixANT architecture consists of a sequence of state-space model blocks, where we utilize a hybrid approach combining standard Mamba blocks and our proposed MixMamba blocks. This design allows the model to extract basic features in the earlier layers and benefit from the enhanced selective capabilities in the later layers.

The overall architecture processes the input to predict future actions through a sequence of $K$ total blocks. Among these, the first $K_0$ blocks utilize standard Mamba blocks, while the remaining $K_E = K - K_0$ blocks employ our proposed MixMamba blocks, as illustrated in \cref{fig:main}(a). Each block follows a similar structure to transformer blocks, with normalization, state-space processing, and MLP components, as illustrated in \cref{fig:main}(b).

Given pre-extracted observed frame features, we first create the conditioning vector $\mathcal{X}$ as described in \cref{eq:padding}. This vector is then concatenated with the latent variables $\hat{Y}_{t}$ along the channel dimension to form $F_t\in \mathbb{R}^{(P + F)\times(n_c + n_d)}$, where $n_c$ is the number of classes and $n_d$ is the number of feature channels. For the $k^{th}$ block, the processing flow is as follows:
\begin{align}
    \hat{F}_t^{k} = 
    \begin{cases}
        \text{Mamba}(\text{LN}(F_{t}^{k-1})), & \text{if } k \leq K_0 \\
        \text{MixMamba}(\text{LN}(F_{t}^{k-1})), & \text{if } k > K_0,
    \end{cases}
\end{align}
where LN represents Layer Normalization \cite{ba2016layer_norm}. It needs to be emphasized that $\gamma(F_{t,1:P}^{k-1})$ in \cref{eq:gating_vector} 
is only conditioned on the features of the observed frames. Ablation on conditioning MixMamba on both observed and future frames is provided in the suppl.\ material. 
After the state-space processing, an MLP layer is applied, followed by a residual connection:
\begin{align} 
    F_t^{k} = \text{MLP}(\hat{F}_t^{k}) + F_t^{k-1}
\end{align}

After processing through all $K$ blocks, we generate the final prediction $\hat{Y}_{t-1} \in \mathbb{R}^{(P+F) \times n_c}$ using an MLP layer, representing the per-frame action predictions for both observed and future frames.

\subsection{Training}
\label{sec:training_inference}

We train our MixANT model using the established diffusion training approach~\cite{zatsarynna2024gtd} with an additional load balancing mechanism to ensure effective utilization of all expert matrices $\{\textbf{A}_1, \textbf{A}_2, ..., \textbf{A}_E\}$ in our mixture approach.

For each training step, we sample a diffusion step $t$ uniformly at random and obtain the corresponding noisy action sequence $\tilde{Y}_t$ using the forward diffusion process. We then pass $\tilde{Y}_t$ alongside the conditioning vector $\mathcal{X}$ to our MixANT model to obtain a reconstruction $\hat{{Y}}_{0}$ of the ground-truth action sequence ${Y}$. The primary reconstruction loss is the $\mathcal{L}_2$-loss between the predicted and the one-hot-encoded ground-truth action sequences:
\begin{align}
    \mathcal{L}_{rec} = || Y - \hat{Y}_{0}||^2.
\end{align}

To ensure balanced utilization of the expert matrices in our mixture approach, we propose a load-balancing mechanism. During training, we track the usage of each expert by computing:
\begin{align}
    C_e^k = \sum_{b=1}^{B} \gamma_e^k(F_{t,1:P}^{k-1}(b))
    \label{eq:expert_weight}
\end{align}
where $C_e^k$ represents the weight for expert $e$ in layer $k$ in the batch of size $B$. The load balancing loss then encourages uniform expert utilization by minimizing:
\begin{align}
    \mathcal{L}_{lb} = \sum_{k=K_0+1}^{K} \text{KL}\bigg(\frac{C^k}{\sum_e C_e^k} \Big\|\, \mathcal{U}(E) \bigg)
\end{align}
where $\mathcal{U}(E)$ represents a uniform distribution over $E$ experts, and KL denotes the Kullback-Leibler divergence.

Our combined training objective is then given by
\begin{align}\label{eq:loss}
    \mathcal{L}_{total} = (1-\lambda_{lb} )\mathcal{L}_{rec} + \lambda_{lb} \cdot \mathcal{L}_{lb},
\end{align}
where $\lambda_{lb}$ controls the contribution of the load balancing loss. The impact of $\lambda_{lb}$ is evaluated in the suppl.\ material.
\section{Experiments}

\textbf{Evaluation Settings:}
We evaluate our proposed MixANT model using the stochastic long-term dense action anticipation protocol introduced by Farha \textit{et al.}~\cite{farha2019uaaa}. In this protocol, the observation length and anticipation horizon are specified as proportions of the total video duration. For a video \( v \) with \( n_v \) frames, the model observes the first \( P = \alpha n_v \) frames and predicts action labels for the subsequent \( F = \beta n_v \) frames, where \( \alpha \) and \( \beta \) represent the observation and anticipation ratios.
%
For each observed video segment, we generate $S{=}25$ prediction samples from our model. We report two key metrics: Mean MoC (Mean over Classes) accuracy and Top-1 MoC accuracy. Mean MoC measures the average accuracy across all $S$ generated samples, while Top-1 MoC reflects the accuracy of the best-matching sample. Consistent with previous works, we evaluate with observation ratios $\alpha{\in}\{0.2, 0.3\}$ and anticipation horizons $\beta{\in}\{0.1, 0.2, 0.3, 0.5\}$. Our experiments utilize three standard datasets for action anticipation:

\textit{Breakfast}~\cite{Kuehne2014_breakfast} comprises 1,712 cooking videos showing 10 different breakfast meal preparations across various kitchen environments. The videos are annotated with 48 action classes and vary in length, with the longest sequence requiring anticipation of up to 5.4 minutes. We evaluate using the established 4-fold cross-validation and report average performance across these splits.

\textit{Assembly101}~\cite{sener2022_Assembly101} is a large-scale dataset containing 4,321 videos of toy assembly tasks from multiple viewpoints. The videos are densely annotated with 202 coarse action classes, with the longest sequences extending beyond 25 minutes and requiring anticipation up to 12.5 minutes into the future. Following standard protocol, we train on the training set and evaluate on the validation set.

\textit{50Salads}~\cite{Stein2013_50Salads} consists of 50 videos showing salad preparation, annotated with 17 action classes. The mean video duration is 6.4 minutes, with the longest sequence requiring anticipation of up to 5.1 minutes. We report results averaged over the 5 standard cross-validation splits.

For a fair comparison with previous methods, we use identical pre-extracted frame features. Specifically, we utilize I3D features from~\cite{farha2020gcpr} and \cite{gong2022future} for Breakfast and 50Salads datasets, while for Assembly101 we employ TSM-features~\cite{Lin2020TSMTS} provided by~\cite{sener2022_Assembly101}.

\noindent\textbf{Implementation Details:}
Our MixANT architecture consists of $K=15$ sequential processing blocks. To balance computational efficiency with the benefits of our mixture approach, we implement a hybrid design where the first $K_0=3$ blocks use standard bidirectional Mamba operations without mixtures, while the remaining $K_E=12$ blocks incorporate our proposed mixture mechanism with $E=5$ experts. This configuration allows the model to establish foundational representations in the initial layers while leveraging the enhanced selective capabilities of the mixture approach in the deeper layers. Additional implementation details are provided in the suppl.\ material.

\definecolor{tablecolor}{rgb}{0.969, 0.969, 0.71}

\begin{table}
\centering
\arrayrulecolor{gray}
\resizebox{\linewidth}{!}{%
\begin{tabular}{ll rrrr | rrrr }
\toprule

\multirow{2}{*}{MoC} & \multirow{2}{*}{Method} & \multicolumn{4}{c}{$\beta \hspace{2pt} (\alpha=0.2)$} & \multicolumn{4}{c}{$\beta \hspace{2pt} (\alpha=0.3) $} \\
\cline{3-10}
& & \textit{0.1} & \textit{0.2} & \textit{0.3} & \textit{0.5} & \textit{0.1} & \textit{0.2} & \textit{0.3} & \textit{0.5} \\

\midrule
\midrule

\multicolumn{10}{c}{Breakfast} \\
\midrule

\multirow{5}{*}{\makecell{Mean}} 

& Tri-gram~\cite{farha2019uaaa}
& 15.4 & 13.7 & 12.9 & 11.9
& 19.3 & 16.6 & 15.8 & 13.9 \\

& UAAA~\cite{farha2019uaaa}
& 15.7 & 14.0 & 13.3 & 13.0
& 19.1 & 17.2 & 17.4 & 15.0 \\

& DiffAnt~\cite{zhong2023diffant}
& 24.7 & 22.9 & 22.1 & 22.3
& 30.9 & 30.2 & 28.9 & 27.5 \\

& GTDA ~\cite{zatsarynna2024gtd}
& 24.0 & 22.0 & 21.4 & 20.6
& 29.1 & 26.8 & 25.3 & 24.2 \\

& MANTA~\cite{zatsarynna2025manta}
& 27.7 & 25.3 & 24.6 & 23.8
& 34.2 & 30.9 & 29.1 & 27.7 \\

\rowcolor{tablecolor}
\cellcolor{white} & \textbf{MixANT}
& \textbf{29.6} & \textbf{26.3} & \textbf{25.9} & \textbf{25}
& \textbf{36.2} & \textbf{32.8} & \textbf{31.2} & \textbf{28.7} \\

\midrule

\multirow{5}{*}{\makecell{Top-1}} 

& Tri-gram~\cite{farha2019uaaa}
& - & - & - & -
& - & - & - & - \\

& UAAA~\cite{farha2019uaaa}
& 28.9 & 28.4 & 27.6 & 28.0
& 32.4 & 31.6 & 32.8 & 30.8 \\

& DiffAnt~\cite{zhong2023diffant}
& 31.3 & 29.8 & 29.4 & 30.1
& 37.4 & 37.0 & 36.3 & 34.8 \\

& GTDA~\cite{zatsarynna2024gtd}
& 51.2 & 47.3 & 45.6 & 45.0
& 54.0 & 50.4 & 49.6 & 47.8 \\

& MANTA~\cite{zatsarynna2025manta}
& {55.5} & {51.0} & {47.9} & {46.9}
& 59.6 & 55.0 & 53.7 & 51.9 \\

\rowcolor{tablecolor}
\cellcolor{white} & \textbf{MixANT}
& \textbf{57.1} & \textbf{52.0} & \textbf{49.1} & \textbf{48.4}
& \textbf{60.7} & \textbf{56.3} & \textbf{55.5} & \textbf{53.5} \\

\midrule
\multicolumn{10}{c}{Assembly101} \\
\midrule

\multirow{4}{*}{\makecell{Mean}}

& Tri-gram~\cite{zatsarynna2024gtd}
& 2.8 & 2.2 & 1.9 & 1.5 
& 3.5 & 2.7 & 2.3 & 1.8 \\

& UAAA~\cite{farha2019uaaa}
& 2.7 & 2.1 & 1.9 & 1.7
& 2.4 & 2.1 & 1.9 & 1.7 \\

& GTDA~\cite{zatsarynna2024gtd}
& 6.4 & 4.5 & 3.5 & 2.8 
& 5.9 & 4.2 & 3.5 & 2.9 \\

& MANTA~\cite{zatsarynna2025manta}
& 6.7 & 5.3 & 4.2 & 3.5
& 6.6 & 4.7 & 4.2 & 3.5 \\

\rowcolor{tablecolor}
\cellcolor{white} & \textbf{MixANT}
& \textbf{8.0} & \textbf{7.0} & \textbf{6.2} & \textbf{4.4}
& \textbf{8.7} & \textbf{6.6} & \textbf{5.6} & \textbf{4.6} \\

\midrule

\multirow{4}{*}{\makecell{Top-1}}

& Tri-gram~\cite{zatsarynna2024gtd}
& 9.0 & 8.0 & 7.2 & 5.6
& 9.5 & 8.2 & 7.8 & 5.9 \\

& UAAA*~\cite{farha2019uaaa}
& 6.9 & 5.9 & 5.6 & 5.1
& 5.9 & 5.5 & 5.2 & 4.9 \\

& GTDA~\cite{zatsarynna2024gtd}
& 18.0 & 12.8 & 9.9 & 7.7
& 16.0 & 11.9 & 10.2 & 7.7 \\

& MANTA~\cite{zatsarynna2025manta}
& 16.9 & 13.3 & 10.2 & 8.8
& 15.6 & 12.0 & 11.1 & 8.4 \\

\rowcolor{tablecolor}
\cellcolor{white} & \textbf{MixANT}
& \textbf{20.3} & \textbf{14.7} & \textbf{11.3} & \textbf{9.8}
& \textbf{18.2} & \textbf{13.8} & \textbf{13.1} & \textbf{10.2} \\

\midrule
\multicolumn{10}{c}{50Salads} \\
\midrule

\multirow{4}{*}{\makecell{Mean}}

& Tri-gram~\cite{farha2019uaaa}
& 21.4 & 16.4 & 13.3 & 9.4
& 24.6 & 15.6 & 11.7 & 8.6 \\

& UAAA~\cite{farha2019uaaa}
& 23.6 & 19.5 & 18.0 & 12.8
& 28.0 & 18.0 & 14.8 & 12.1 \\

& GTDA~\cite{zatsarynna2024gtd}
& 28.3 & 22.1 & 17.8 & 11.7
& 29.9 & 18.5 & 14.2 & 10.6 \\

& MANTA~\cite{zatsarynna2025manta}
& 28.6 & 22.8 & 19.5 & 13.6
& 31.3 & 21.9 & 17.6 & 13.0 \\

\rowcolor{tablecolor}
\cellcolor{white} & \textbf{MixANT}
& \textbf{30.3} & \textbf{25.0} & \textbf{20.9} & \textbf{15.2}
& \textbf{33.4} & \textbf{23.7} & \textbf{19.7} & \textbf{14.6} \\

\midrule

\multirow{4}{*}{\cellcolor{white} \makecell{Top-1}} 

& Tri-gram~\cite{farha2019uaaa}
& - & - & - & -
& - & - & - & - \\

& UAAA*~\cite{farha2019uaaa}
& 53.5 & 43.0 & 40.5 & 33.7
& 56.4 & 42.8 & 35.8 & 30.2 \\

& GTDA~\cite{zatsarynna2024gtd}
& 69.6 & 55.8 & 45.2 & 28.1
& 66.2 & 44.9 & 39.2 & 31.0 \\

& MANTA~\cite{zatsarynna2025manta}
& 68.3 & 51.5 & 41.7 & 31.3
& 71.7 & 53.3 & 43.8 & 31.1 \\

\rowcolor{tablecolor}
\cellcolor{white} & \textbf{MixANT}
& \textbf{71.5} & \textbf{56.9} & \textbf{46.5} & \textbf{35.0}
& \textbf{72.9} & \textbf{54.6} & \textbf{44.9} & \textbf{32.4} \\

\midrule

\end{tabular}
}
\vspace{-0.2cm}
\caption{Comparison of MixANT to state-of-the-art methods on Breakfast, Assembly101, and 50Salads.}
\vspace{-0.4cm}
\label{tab:results_main}
\end{table}

\subsection{Comparison to State of the Art}

\cref{tab:results_main} presents a comparison of our MixANT approach against previous methods across all three datasets. Our model consistently outperforms existing techniques, establishing new state-of-the-art results across virtually all evaluation settings and metrics.

On the Breakfast dataset, MixANT demonstrates substantial improvements over prior approaches, including MANTA~\cite{zatsarynna2025manta} and GTDA~\cite{zatsarynna2024gtd}. For both Mean MoC and Top-1 MoC metrics, our method shows consistent gains across all anticipation horizons. Notably, for Mean MoC with $\alpha{=}0.3$ and $\beta{=}0.1$, MixANT achieves 36.2\%, compared to MANTA's 34.2\%, while for Top-1 MoC, we achieve 60.7\% versus 59.6\%. The challenging Assembly101 dataset, with its 202 action classes, presents a more complex anticipation scenario where MixANT demonstrates even more substantial improvements. For Mean MoC, our approach shows relative gains of up to 32\% over MANTA at longer anticipation horizons. For Top-1 MoC, MixANT surpasses the previous state-of-the-art GTDA in most settings, achieving 20.3\% with $\alpha{=}0.2$ and $\beta{=}0.1$.

On the 50Salads dataset, MixANT again establishes new state-of-the-art results. Our method achieves the highest performance for both Mean MoC across all settings and Top-1 MoC for nearly all configurations, with particularly strong performance at shorter anticipation horizons. These consistent improvements across diverse datasets demonstrate the effectiveness of our mixture-based approach for the state-transition matrix. By enabling adaptive temporal memory processing through expert specialization, MixANT significantly enhances the model's capacity to generate accurate and contextually relevant action predictions across varying time scales.

\subsection{Analysis}
In this section, we discuss a series of analysis and ablation experiments that we performed to explore individual components of our proposed MixANT model. All experiments use the Breakfast dataset with $\alpha{=}0.2$ and $\beta{=}0.1$. 

\begin{figure}[t]
    \centering
    \includegraphics[width=\columnwidth]{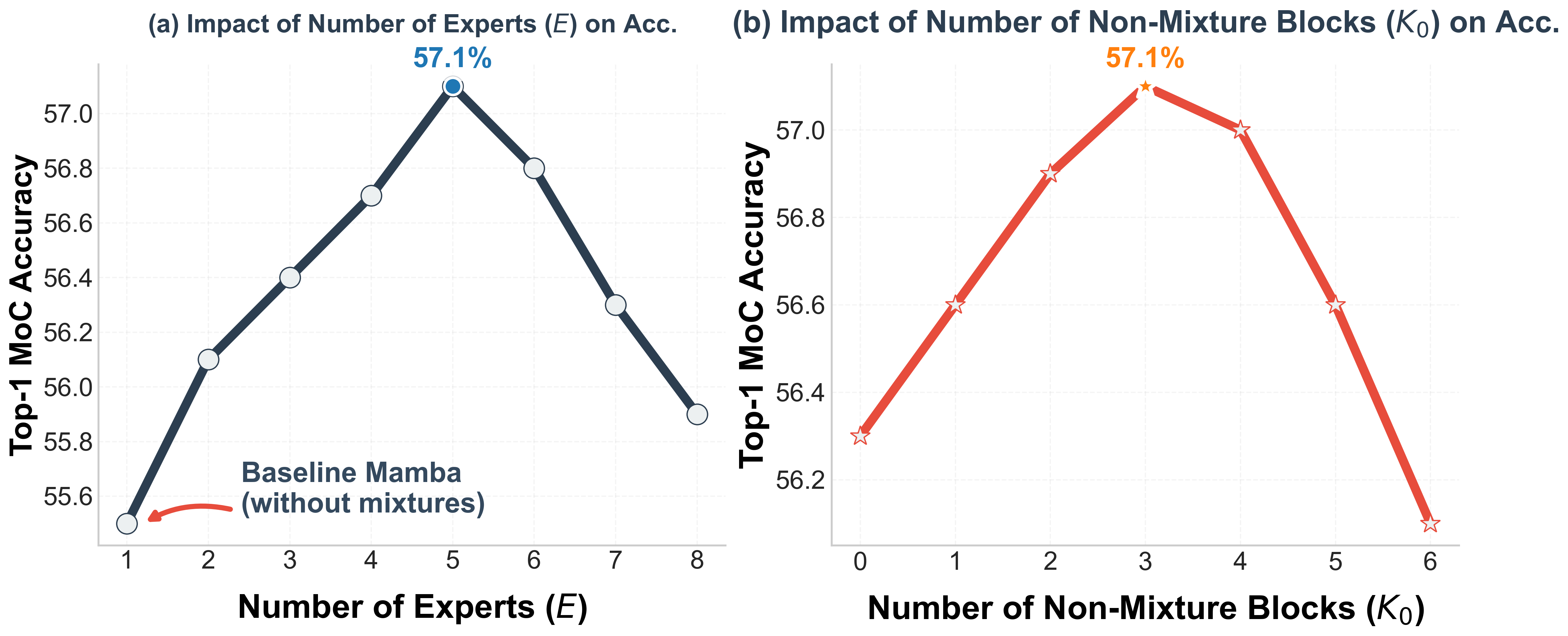}
    \caption{(a) Impact of the number of experts in each layer. (b) Effect of the number of initial static blocks ($K_0$).}
    \label{fig:experts_initblocks_ablation}
    \vspace{-2mm}
\end{figure}

\noindent\textbf{Number of experts per block:}
A design choice in our work is the number of experts per block. MixANT allows for flexibility in this parameter, starting from a single expert (equivalent to a standard Mamba block) and scaling up as needed. We conducted a systematic experiment, increasing the number of experts from 1 to 8.
As shown in \cref{fig:experts_initblocks_ablation}(a), the model performance improves significantly as we increase the number of experts from the baseline (\ie 1 expert), reaching optimal results at approximately 5 experts per block. Beyond this point, we observe a performance decline as additional experts are introduced.
This pattern reveals important insights about expert-based architectures. The initial performance gains can be attributed to two mechanisms: (1) increased model capacity and expressiveness through parallel processing pathways, and (2) expert specialization across diverse input patterns. Each expert effectively learns to handle a distinct subset of the data distribution, collectively enabling more nuanced representations than a single expert.
However, the performance degradation beyond 5 experts highlights a fundamental training challenge in mixture-of-experts architectures. With a fixed training budget, models with excessive experts suffer from sparse gradient updates. Each expert receives fewer training signals, creating an effective data starvation that impedes further gains in performance. This observation suggests that the scaling of the expert count should be proportional to available training data and computational resources.

\noindent\textbf{Number of initial static blocks:}
Another design parameter in our architecture is the number of initial static blocks. These blocks implement standard Mamba processing with a single expert, before transitioning to our proposed mixture-of-experts configuration. To determine the optimal setup, we systematically investigated the number of initial static blocks from 0 to 6, as shown in \cref{fig:experts_initblocks_ablation}(b).
Our results reveal a clear pattern: performance improves significantly as we increase the number of initial static blocks up to 3, after which we observe a consistent decline. This finding demonstrates that expert routing is most effective when applied to intermediate and higher-level representations rather than raw input features.
The information processing hierarchy within the network can explain this behavior. Early layers typically extract general, low-level features that benefit from consistent processing across all inputs. Introducing input-dependent routing too early forces premature specialization before the model has extracted meaningful features to guide expert selection. The initial static blocks provide a foundation of general feature extraction, allowing subsequent expert layers to make more informed routing decisions based on these refined representations.
Conversely, having too many static blocks (beyond 3) constrains the network's expressiveness by limiting the layers that can benefit from adaptive, input-dependent processing. Each expert block adds capacity and flexibility to the model, so reducing their number effectively caps the model's ability to specialize across diverse input patterns. 
These results also highlight the importance of making the $\textbf{A}$ matrix input-dependent since not introducing enough $\textbf{A}$ matrices or experts hinders the model's performance.

\begin{figure}[t]
    \centering
    \includegraphics[width=\columnwidth]{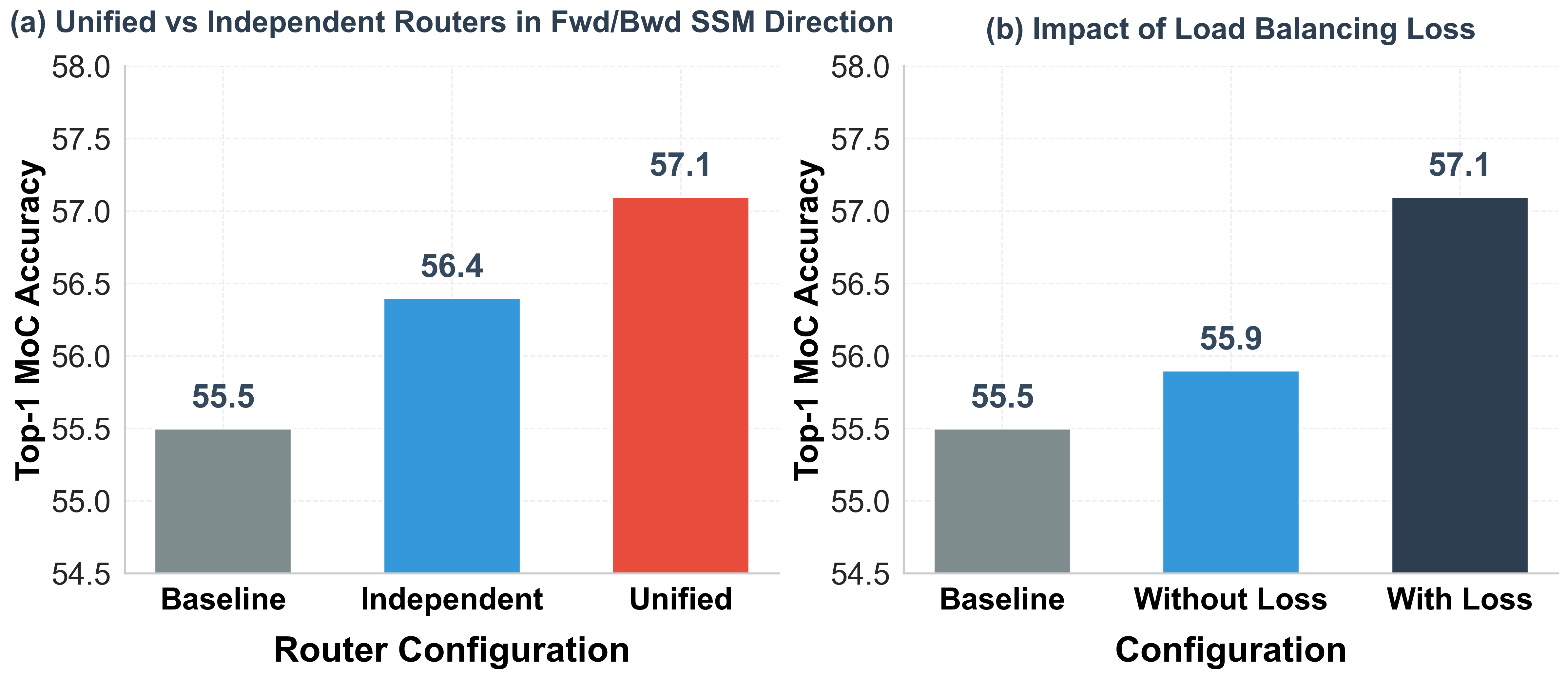}
    \vspace{-6mm}
    \caption{(a) Comparison of router configurations: unified (same experts in forward and backward passes) versus independent (different experts selected in forward and backward passes). (b) Impact of load balancing loss. }
    \label{fig:loss_router_ablation}
    \vspace{-3mm}
\end{figure}

\noindent\textbf{Router configuration:}
In the MixMamba layer (\cref{fig:main}(c)), there are two MixSSM units. One processes the input in the forward direction and the other one in the backward direction. Each of these MixSSM units has its own mixture of $\textbf{A}$ matrices. In this case, the routing can be done separately (independent case) or collectively for both units (unified case). In separate routing, there are two learnable projection matrices $W_{g}^1$ and $W_{g}^2$ from which two gating vectors $\gamma^1$ and $\gamma^2$ are computed. In this case, the $\textbf{A}$ matrix chosen for the forward and backward units depends on the index of the maximum values of $\gamma^1$ and $\gamma^2$ after the softmax operation, respectively. In the unified case, we compute only one gating vector $\gamma$, which is then used for both forward and backward units. Our results in~\cref{fig:loss_router_ablation}(b) show that the unified router configuration performs better than the independent one. This is intuitive since in an independent configuration, bi-directionality of the SSM is sacrificed because the forward and backward units are learning two different $\textbf{A}$ matrices instead of learning the $\overrightarrow{A}$ matrix for the forward direction and its counterpart $\overleftarrow{A}$ for the backward direction.  

\begin{figure}[t]
   \centering
   \includegraphics[width=0.8\columnwidth]{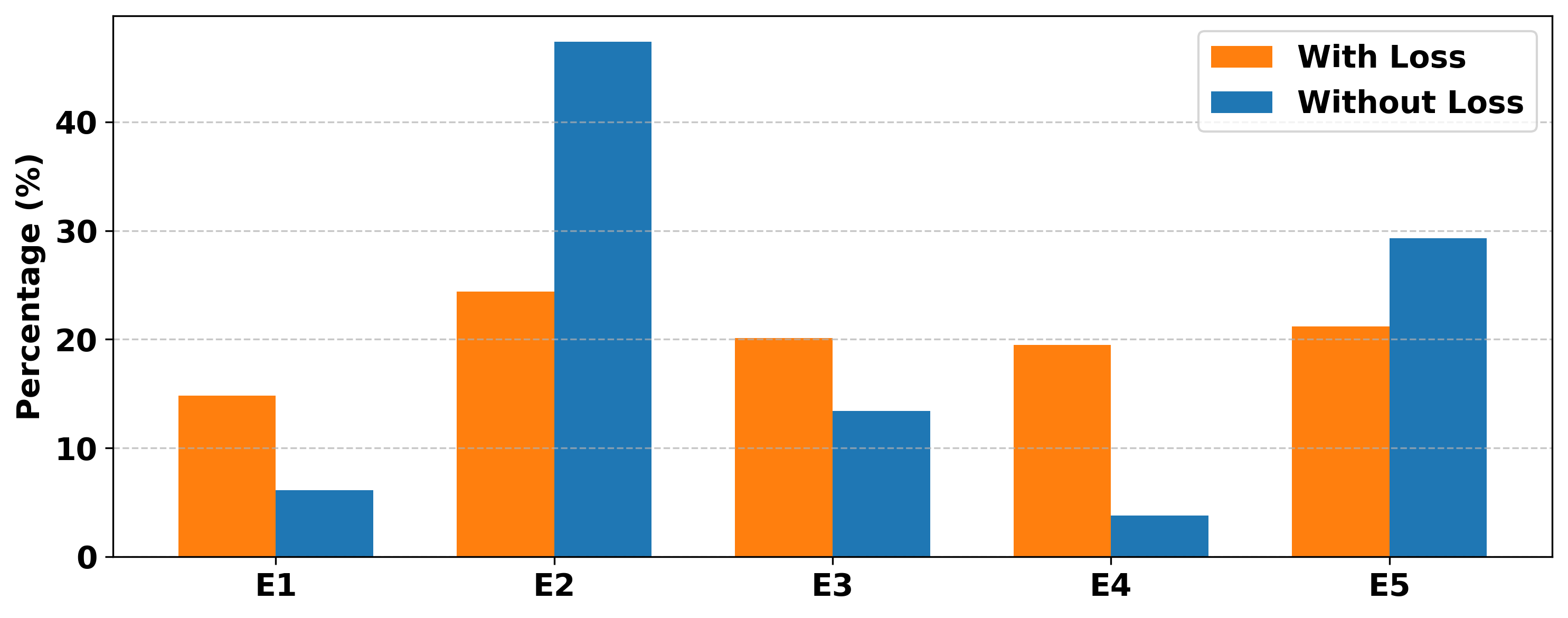}
   \vspace{-2mm}
   \caption{Utilization of the 5 experts across all layers on Breakfast.}
   \label{fig:expert_usage}
   \vspace{-3mm}
\end{figure}

\noindent\textbf{Load balancing and expert usage:}
The load balancing mechanism explicitly encourages uniform utilization of all experts during training as detailed in \cref{sec:training_inference}. This constraint is removed during inference. As demonstrated in \cref{fig:loss_router_ablation}(b), even without load balancing, our expert-based architecture outperforms the baseline due to its expanded representational capacity. However, these gains are significantly enhanced when load balancing is applied, as it ensures all experts receive adequate training signals to develop specialized functionality, as shown in \cref{fig:expert_usage}. Without load balancing, the second expert is selected in nearly 50\% of the cases, while E1 and E4 are barely selected. The optimal utilization of the available expert capacity with load-balancing translates directly to improved performance metrics, validating the importance of balanced expert training in the MixANT architecture.

\noindent\textbf{Additional ablations:}
In the suppl.\ material, we provide additional ablation experiments regarding the conditioning of the gating vector $\gamma$ in \eqref{eq:expert_weight}, impact of $\lambda_{lb}$ in \eqref{eq:loss}, and a comparison of the proposed mixture-of-experts approach to a query-key approach or a very large MLP for making $\textbf{A}$ input-dependent.

\subsection{MixANT's Expert Selection Patterns}
\begin{figure}[t]
    \centering
    \includegraphics[width=0.95\columnwidth]{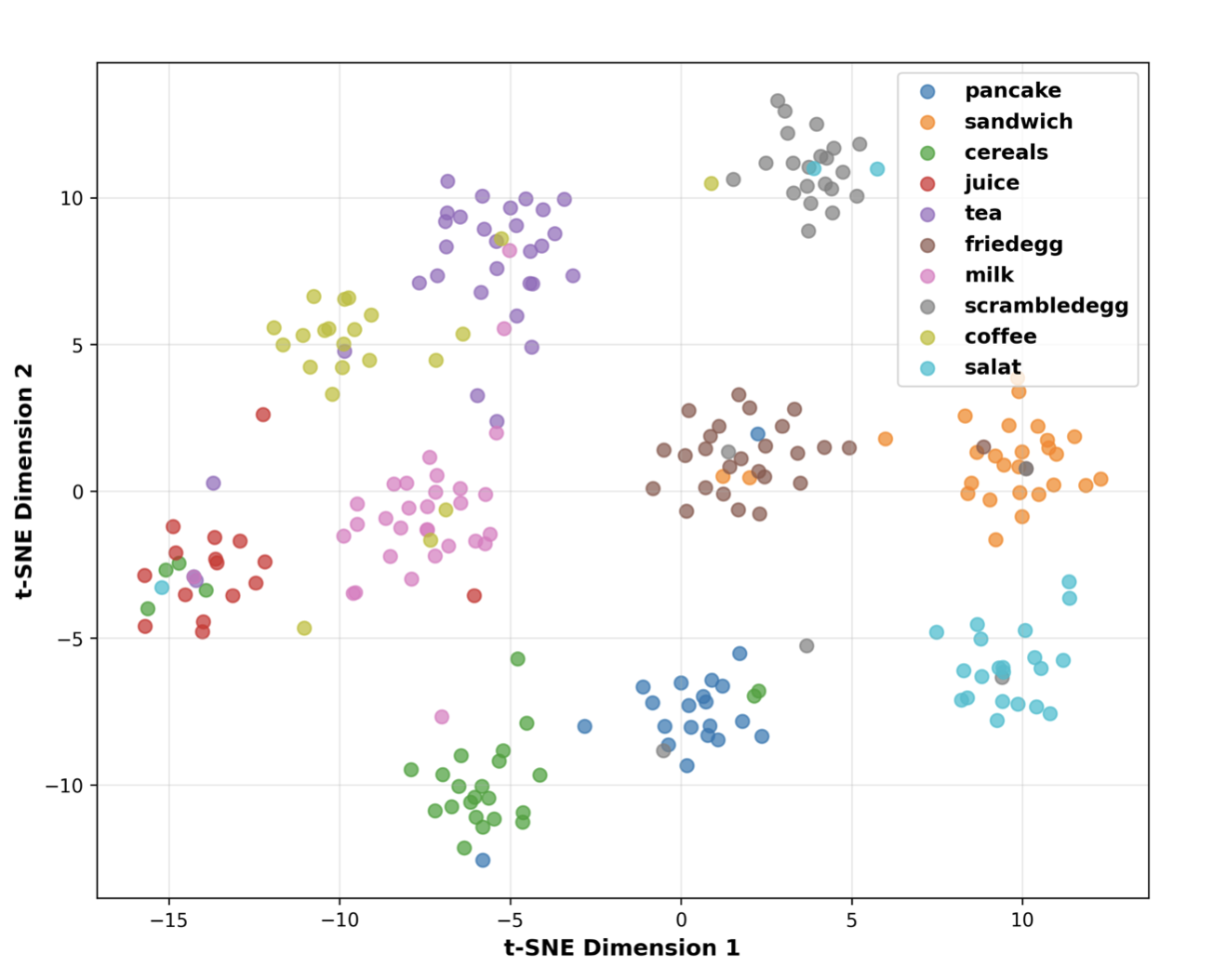}
\captionsetup{width=\columnwidth}
    \vspace{-2mm}
    \caption{The t-SNE plot shows that using only a binary vector of the selected experts for each input sequence separates high-level activities. This shows that the selection of the experts in the MixANT model depends on the semantic context of the video.}
    \label{fig:tsne}
    \vspace{-2mm}
\end{figure}

To understand MixANT's operational dynamics, we analyze expert selection patterns using the test split 1 of Breakfast with $\alpha{=}0.2$ and $\beta{=}0.1$. For each testing instance, we initialize a selection matrix $S\in \mathbb{R}^{(K_{E}\times E)}$, where $K_{E}$ represents the total number of mixture blocks and $E$ is the number of experts per block. As input sequences propagate through the network, we track which expert $\mathbf{A}$ matrix is selected at each block. For each block row in $S$, we set a value of 1 in the column corresponding to the chosen expert while other columns remain zero. After processing each instance, we flatten these matrices and visualize them using t-SNE dimensionality reduction in \cref{fig:tsne}.

The results reveal a striking emergent property: despite being trained only on atomic actions (cutting, stirring, pouring) rather than complete activity categories (making salad, preparing tea), MixANT implicitly learns to contextualize these actions within broader activities. The t-SNE visualization shows a clear clustering of expert selection patterns according to high-level activity categories, even without explicit training on these umbrella categories. This organization indicates that MixANT's expert matrices develop specialized functions aligned with semantic activity structures, enabling more nuanced temporal reasoning.

\section{Conclusion}

In this work, we presented MixANT, a novel architecture that enhances Mamba-based models for stochastic long-term dense action anticipation by introducing input-dependent selectivity to the forget-gate ($\textbf{A}$ matrix). Our mixture of experts approach dynamically selects the most contextually relevant $\textbf{A}$ matrix based on input features, providing crucial temporal memory control without sacrificing computational efficiency. Extensive experiments on the 50Salads, Breakfast, and Assembly101 datasets demonstrate that MixANT consistently outperforms existing state-of-the-art methods across all evaluation settings. These results confirm our hypothesis that input-dependent control of the hidden state evolution is particularly beneficial for anticipation tasks where the relevance of past information varies greatly across contexts.
\section*{Acknowledgement}
The work has been supported by the ERC Consolidator Grant FORHUE (101044724), the Federal Ministry of Education and Research (BMBF) under grant no. 01IS22094A WEST-AI, the Deutsche Forschungsgemeinschaft (DFG, German Research Foundation) GA 1927/4-2 (FOR 2535 Anticipating Human Behavior), and the project iBehave (receiving funding from the programme ``Netzwerke 2021”, an initiative of the Ministry of Culture and Science of the State of North Rhine-Westphalia). For the computations involved in this research, we acknowledge EuroHPC Joint Undertaking for awarding us access to Leonardo at CINECA, Italy, through EuroHPC Regular Access Call - proposal No.\ EHPC-REG-2025R01-218.
{
    \small
    \bibliographystyle{ieeenat_fullname}
    \bibliography{main}
}
\clearpage
\clearpage
\setcounter{page}{1}
\maketitlesupplementary

\section{Introduction}
\label{sec:supp_intro}
This supplementary material provides comprehensive details regarding the implementation, ablation studies, and qualitative analysis of our proposed MixANT model. The document is structured as follows: Section~\ref{sec:supp_impl} describes the implementation details of our model architecture; Section~\ref{sec:supp_abl} presents additional ablation studies to analyze the contribution of different components; Section~\ref{sec:supp_ant} provides some additional results on other anticipation tasks; and Section~\ref{sec:supp_qual} offers further qualitative comparisons between MixANT and existing approaches in the literature.

\begin{figure}[b]
    \centering
    \includegraphics[width=\columnwidth]{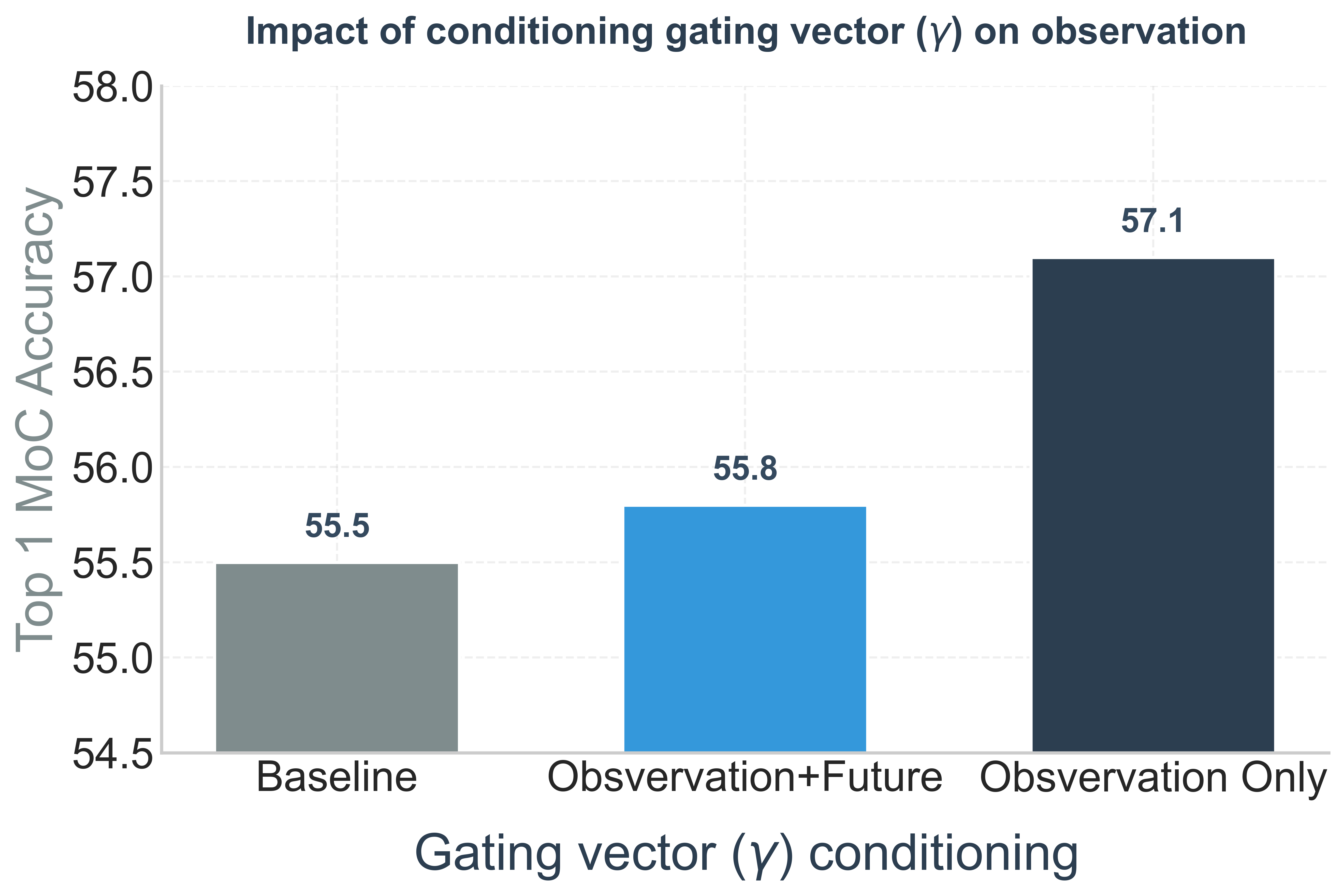}
    \caption{Impact of conditioning gating vector $\gamma$ on present and future.}
    \label{fig:gating_vector_conditioning}
\end{figure}

\section{Implementation Details}
\label{sec:supp_impl}
\begin{figure*}[t]
    \centering
    \includegraphics[width=\textwidth]{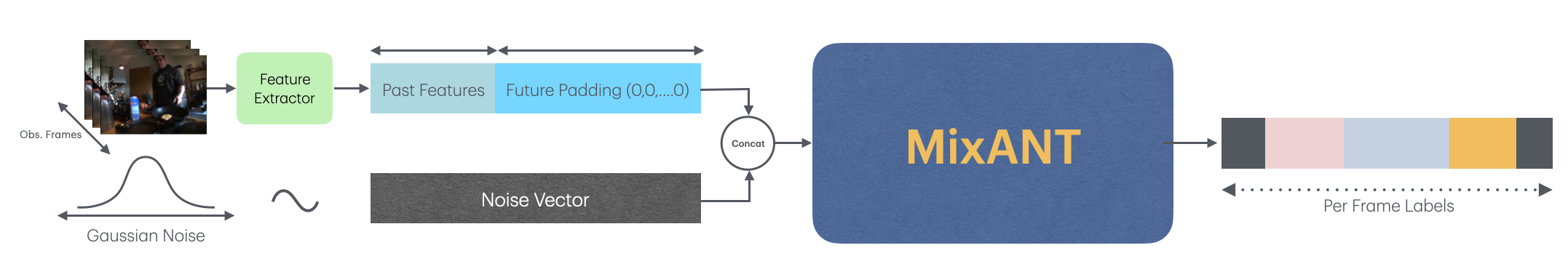}
    \\
    \captionsetup{width=\textwidth}
    \caption{Overall task of stochastic long-term dense action anticipation with its inputs and outputs. For the Breakfast and 50Salads datasets, we extract I3D features from previous works~\cite{farha2020gcpr} and \cite{gong2022future}, while for Assembly101 we use TSM-features~\cite{Lin2020TSMTS} provided by~\cite{sener2022_Assembly101}. The features are then concatenated with zero padding to represent future frames. A Gaussian noise vector is sampled and added to the input tensor, which is then processed by the MixANT model to generate per-frame labels. Importantly, each noise vector produces a single sample, and multiple outputs are generated by repeating the process with different noise vectors while maintaining the same input sample.}
    \label{fig:overall_task}
\end{figure*}

The overall task for the long-term dense anticipation is shown in ~\cref{fig:overall_task}. For the Breakfast and 50Salads datasets, we extract I3D features from previous works~\cite{farha2020gcpr} and \cite{gong2022future}, while for Assembly101 we use TSM-features~\cite{Lin2020TSMTS} provided by~\cite{sener2022_Assembly101}. The features are then concatenated with zero padding to represent future feature frames. A Gaussian noise vector is sampled and added to the input tensor, which is then processed by the MixANT model to generate per-frame labels. Importantly, each noise vector produces a single sample, and multiple outputs are generated by repeating the process with different noise vectors while maintaining the same input sample. 

The list of hyperparameters for reproducibility purposes is provided in \cref{tab:hyperparameters}. We used a single A100 GPU (80 GB) for training all of our MixANT models.

\begin{figure}[t]
    \centering
    \includegraphics[width=\columnwidth]{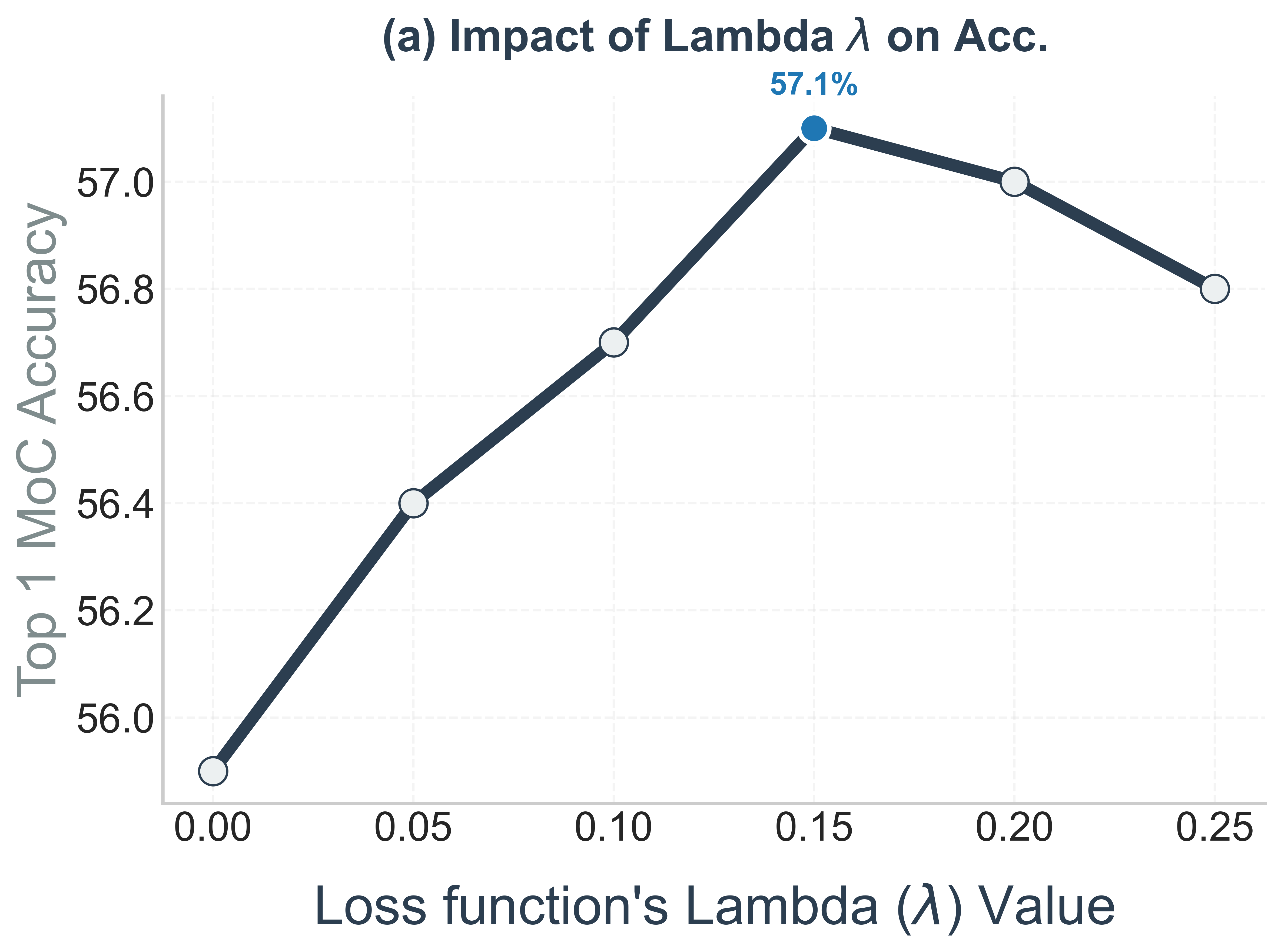}
    \caption{Impact of $\lambda_{lb}$ for the load balancing loss ($\mathcal{L}_{lb}$). }
    \label{fig:impact_of_lambda}
\end{figure}

\begin{table*}[tb]
    \centering
    \resizebox{1\textwidth}{!}{%
    \begin{tabular}{lcccc}
        \toprule
        \textbf{MixANT Training Recipe} \\
        \midrule
        Dataset & Breakfast &  50Salads & Assembly101\\
        \midrule
        Epochs    & 90 & 90   &  90  \\
        Num of MixANT Blocks  & 15 & 15 & 15 \\
        Optimizer & AdamW & AdamW  & AdamW \\
        Optimizer momentum & $\beta_1=0.9,\beta_2=0.999$ & $\beta_1=0.9,\beta_2=0.999$ & $\beta_1=0.9,\beta_2=0.999$\\
        Learning rate       & 0.0005 & 0.001 & 0.0005 \\
        Diffusion Steps (Training) & 1000 & 1000 & 1000 \\
        DDIM steps & 50 & 10 & 50 \\
        \bottomrule
    \end{tabular}}
    \caption{Hyperparameters for MixANT.}
    \label{tab:hyperparameters}
\end{table*}

\section{Additional Ablation Studies}
\label{sec:supp_abl}
\subsection{Impact of Conditioning Gating Vector on Present and Future}
We conducted an ablation study examining how performance is affected when the gating vector $\gamma$ is conditioned on either the observed part of the input alone or the complete input containing both observed and future parts.

For our ablation, we condition the gating vector on the observed part of the input $F_{t,1:P}^{k-1}$ and the complete input $F_{t,1:P+F}^{k-1}$ as in~\cref{eq:expert_weight}. As shown in \cref{fig:gating_vector_conditioning}, conditioning $\gamma$ exclusively on observed frames yields superior performance compared to conditioning on both observation and future components. This finding is consistent with theoretical expectations, as the latter approach incorporates zero-padded future values that provide no meaningful information for the decision-making process of selecting appropriate $\textbf{A}$ matrices. Consequently, restricting the conditioning of the gating vector to only the observation component—which contains the complete contextual information relevant to the selection process—proves to be more effective while appropriately disregarding uninformative future padding.

\subsection{Impact of Contribution of Load Balancing Loss}
We analyze the impact of incorporating a load-balancing loss component into the overall training loss function. This addition of load balancing loss is controlled by a coefficient $\lambda_{lb}$ that determines the relative contribution of the load balancing loss. To analyze its impact, we systematically varied $\lambda_{lb}$ from 0 to 0.25 in increments of 0.05 in \cref{fig:impact_of_lambda}.

When $\lambda_{lb} = 0$, effectively training without load balancing loss, the network performs marginally better than the baseline Mamba network. However, the introduction of load balancing loss substantially improves performance. The optimal results are achieved at $\lambda_{lb} = 0.15$, beyond which performance degrades as the optimization objective becomes overly focused on load balancing at the expense of overall performance metrics.

\begin{figure}[t]
    \raggedleft
    \includegraphics[width=\columnwidth]{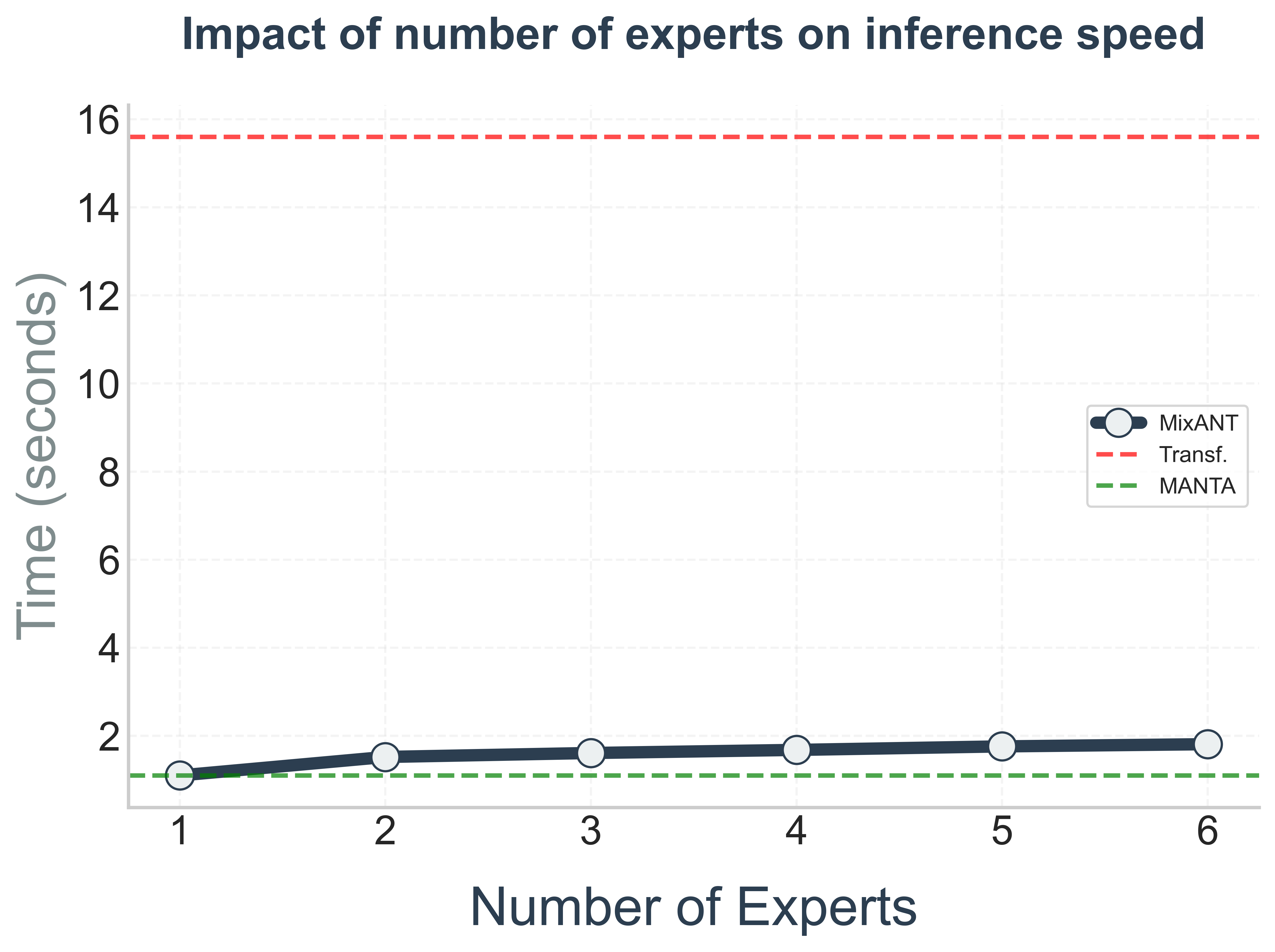}
    \caption{Mean inference time for generating 25 samples using our model on the Breakfast dataset with $\alpha{=}0.3$ and $\beta{=}0.5$ across varying numbers of experts. Dashed horizontal lines represent the inference speed of other methods. }
    \label{fig:inference_speed}
\end{figure}

\subsection{Impact of Number of Experts on Inference Speed}
We evaluate the mean time required for generating 25 samples using our model on the Breakfast dataset with $\alpha{=}0.3$ and $\beta{=}0.5$, for different numbers of experts. The results are presented in \cref{fig:inference_speed}. We observe that there is a slight increase in inference time when the first mixture is used (2 experts). After that increase, the number of experts poses a minimal overhead. We do not include GTDA in the plot because its inference speed is much higher ($71.8$ seconds).

\subsection{Computational Complexity and Performance}
\cref{tab:comp_complexity} compares parameters, memory, and inference time of various methods. We also compare our approach to alternatives to obtain input-dependent A matrices (large MLP and query-key). MixANT is much more efficient than a large MLP ($>80$x faster) and query-key ($>14$x faster), and it achieves a higher Top-1 MoC.

\begin{table}[t]
    \centering
    \resizebox{\columnwidth}{!}{%
    \begin{tabular}{lcccccc}
        \toprule
        Method & Mamba & $A(x)$ & Param. (M) &  Mem. (GB) & Inf. Time (s) & Top-1 MoC \\
        \midrule
        GTDA          &    &     & 3.9                & 19.2  & 71.8  & 48.9 \\
        Transf. (15)   &    &    & 1.2                & 11.3  & 15.6  & 48.8 \\
        Transf. (18)    &    &   & 1.4                & 13.2  & 18.2  & 50.3 \\
        MANTA            &  \checkmark &   & 1.4                & 10.2  & 1.1   & 52.7 \\
        \hline
        Large MLP        &   \checkmark &  \checkmark & 8.0                & 38.7  & 137.4 & 47.4 \\
        Query-Key         &  \checkmark &  \checkmark & 2.3                & 22.7  & 24.4  & 52.3 \\
        MixANT ($E=5$)    &  \checkmark  & \checkmark & 1.6                & 10.9  & 1.7   & 54.1 \\
        \bottomrule
    \end{tabular}}
    \caption{Comparison of computational cost and Top-1 MoC for different methods. The last three rows report results for three variants of making $A$ input-dependent ($A(x)$). Top-1 MoC is averaged over all $\alpha$ and $\beta$ values on Breakfast. Inference time is per video for $\alpha=0.3$, $\beta=0.5$, and generating 25 samples. 
    }
    \label{tab:comp_complexity}
\end{table}

We also report the impact of the number of initial static blocks $K_0$ on parameters, memory, and inference time in~\cref{tab:k_0_computational_comp}.

\begin{table}[t]
    \centering
    \resizebox{1\columnwidth}{!}{%
    \begin{tabular}{lccccccc}
        \toprule
        $K_0$ & 1 & 2 & 3 & 4 & 5 & 6 \\
        \midrule
        Params. (M)   & 1.70 & 1.67 & 1.64 & 1.62 & 1.60 & 1.58 \\
        Mem. (GB)     & 11.0 & 10.9 & 10.9 & 10.8 & 10.8 & 10.7 \\
        Inf. Time (sec) & 1.8 & 1.8 & 1.7 & 1.7 & 1.6 & 1.6 \\
        Top-1         & 53.0 & 53.2 & 53.5 & 53.3 & 52.9 & 52.4 \\
        \bottomrule
    \end{tabular}}
    \vspace{-0.3cm}
    \caption{Computational cost and mean inference time for generating 25 samples using our model on the Breakfast dataset with $\alpha{=}0.3$ and $\beta{=}0.5$ for varying numbers of initial static blocks $K_0$.}
    \label{tab:k_0_computational_comp}
    \vspace{-0.15cm}
\end{table}

\section{Additional Quantitative Results}
\label{sec:supp_ant}

\begin{table*}
\centering
\resizebox{\linewidth}{!}{%
\begin{tabular}{llccccccccc}
\toprule
& & \multicolumn{3}{c}{Overall} & \multicolumn{3}{c}{Unseen} & \multicolumn{3}{c}{Tail} \\
\cmidrule(lr){3-5} \cmidrule(lr){6-8} \cmidrule(lr){9-11}
Method & Block & Verb & Noun & Act & Verb & Noun & Act & Verb & Noun & Act \\
\midrule
Testra~\cite{zhao2022testra} & short Attn~\cite{2024videomambasuite}  & 25.1 & 30.8 & 14.1 & 24.3 & 24.5 & 10.7 & 17.4 & 23.0 & 10.9 \\
Testra~\cite{zhao2022testra} & short Mamba~\cite{2024videomambasuite} & 27.9 & 34.1 & 15.2 & 28.1 & 24.2 & 12.0 & 20.5 & 27.8 & 12.3 \\
\midrule
\rowcolor{gray!20}
Testra~\cite{zhao2022testra} & short MixMamba & 29.7 & 35.6 & 17.1 & 30.4 & 24.8 & 13.5 & 22.7 & 30.2 & 14.1 \\
\bottomrule
\end{tabular}%
}
\caption{Results of action anticipation on EK-100~\cite{damen2022EK100}. Accuracy measured by class-mean recall@5(\%) following the standard protocol. ``short" denotes using short-term memory.}
\label{tab:ek100}
\end{table*}

\cref{tab:ek100} presents the action anticipation results on the EK-100 dataset, comparing our proposed MixMamba approach with traditional attention and Mamba baselines. Our MixMamba method demonstrates consistent improvements across all evaluation metrics and scenarios. Specifically, MixMamba achieves 29.7\% verb accuracy and 17.1\% action accuracy on the overall split, representing improvements of 4.6\% and 3.0\% over the attention baseline, and 1.8\% and 1.9\% over the Mamba baseline, respectively. The improvements are particularly notable in the challenging tail scenarios, where MixMamba reaches 22.7\% verb accuracy and 14.1\% action accuracy, outperforming both baselines by substantial margins. These results demonstrate that our mixture approach is effective on diverse anticipation scenarios.

\section{Additional Qualitative Results}
\label{sec:supp_qual}
\begin{figure*}[t]
    \centering
    \includegraphics[width=\textwidth]{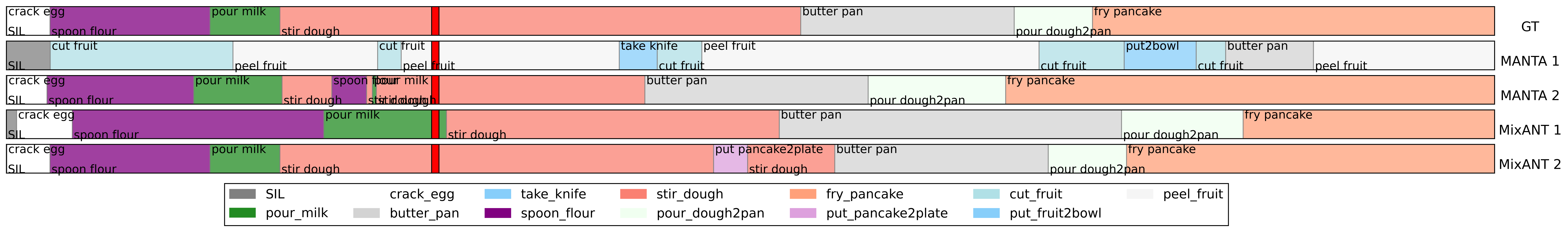}
    \caption{Qualitative figure for anticipation result on the Breakfast dataset for the video P42 ``making a pancake.''}
    \label{fig:qualitative_res_1}
\end{figure*}

\begin{figure*}[t]
    \centering
    \includegraphics[width=\textwidth]{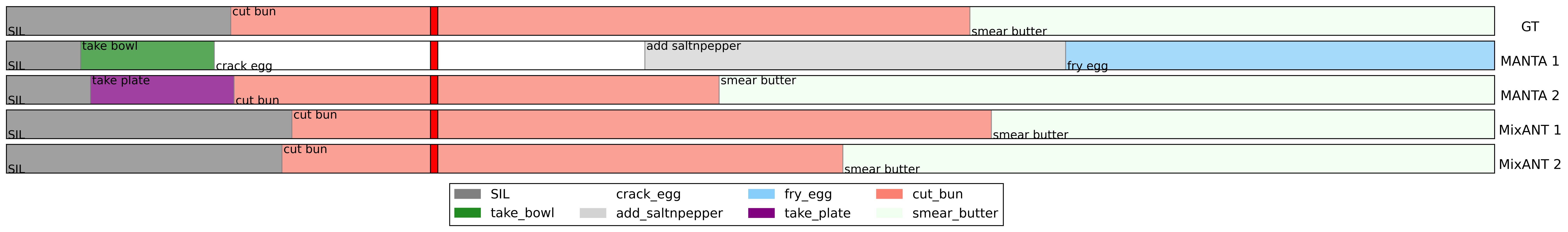}
    \caption{Qualitative figure for anticipation result on the Breakfast dataset for the video P47 ``making a sandwich.''}
    \label{fig:qualitative_res_2}
\end{figure*}

\begin{figure*}[t]
    \centering
    \includegraphics[width=\textwidth]{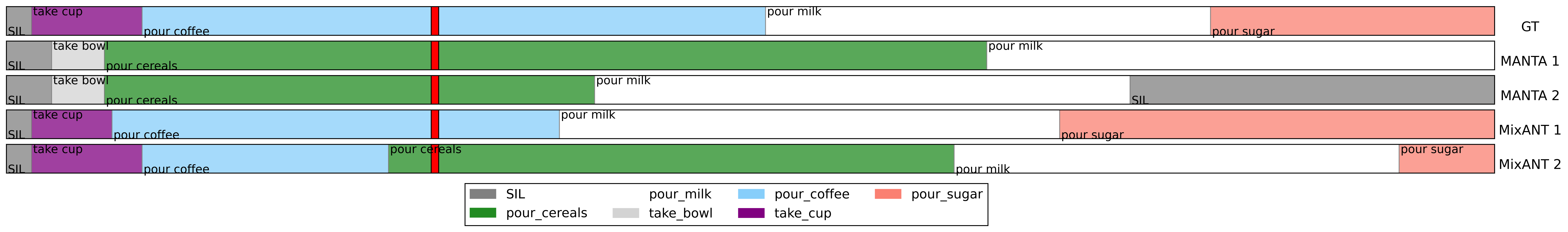}
    \caption{Qualitative figure for anticipation result on the Breakfast dataset for the video P53 ``making coffee.''}
    \label{fig:qualitative_res_3}
\end{figure*}

We present some qualitative results for MixANT in comparison to the baseline MANTA for different action videos on the Breakfast dataset in \cref{fig:qualitative_res_1} (Making Pancake), \cref{fig:qualitative_res_2} (Making Sandwich), and \cref{fig:qualitative_res_3} (Making Coffee). All the qualitative results are for the setting $\alpha{=}0.2$ and $\beta{=}0.5$, and we show two samples for each approach. The results show that our proposed approach is consistently better across all three videos, with greater alignment with the ground truth. 

\end{document}